\documentclass[10pt,twocolumn,letterpaper]{article}

\usepackage[pagenumbers]{cvpr} 

\usepackage[accsupp]{axessibility}
\definecolor{cvprblue}{rgb}{0.21,0.49,0.74}
\usepackage[pagebackref,breaklinks,colorlinks,allcolors=cvprblue]{hyperref}
\usepackage{amsfonts}
\usepackage{amsmath}
\usepackage{array}
\usepackage{multirow}
\usepackage{color}
\usepackage{natbib}
\usepackage{diagbox}
\usepackage{amssymb}
\usepackage{algorithm}
\usepackage{adjustbox}
\usepackage{xspace}
\usepackage{algpseudocode}

\makeatletter
\DeclareRobustCommand\onedot{\futurelet\@let@token\@onedot}
\def\@onedot{\ifx\@let@token.\else.\null\fi\xspace}
\def\eg{\emph{e.g}\onedot} 
\def\ie{\emph{i.e}\onedot}

\makeatother

\DeclareMathOperator*{\argmax}{\arg\max}

\def\paperID{} 
\def\confName{CVPR}
\def\confYear{2026}

\newcommand{\vpara}[1]{\vspace{0.06in}\noindent\textbf{#1 }}
\newcommand{\vvpara}[1]{\vspace{0.06in}\noindent\textit{#1 }}
\newcommand{\model}{\textit{MOON2.0}\xspace}
\newcommand{\bench}{\textit{MBE2.0}\xspace}

\def\one #1 {\textbf{#1}}
\def\two #1 {\underline{#1}~}
\def\thr #1 {*#1}
\def \s   #1 {\footnotesize $\pm$#1}

\title{MOON2.0: Dynamic Modality-balanced Multimodal Representation Learning for E-commerce Product Understanding}

\author{
Zhanheng Nie\footnotemark[1] , Chenghan Fu\footnotemark[1] \ \footnotemark[2] , Daoze Zhang\footnotemark[1] , Junxian Wu\footnotemark[1] , \\ 
Wanxian Guan, Pengjie Wang, Jian Xu, and Bo Zheng\footnotemark[3] \\
Alibaba Group \\
\tt\small \{niezhanheng.nzh,fuchenghan.fch,zhangdaoze.zdz,wujunxian.wjx, \\
\tt\small {wanxian.gwx,pengjie.wpj,xiyu.xj\}@taobao.com,bozheng@alibaba-inc.com}
}

\begin{document}
\maketitle
\renewcommand{\thefootnote}{\fnsymbol{footnote}} 
\footnotetext[1]{Equal Contribution. \quad 
                 \textsuperscript{$\dagger$}Project Leader. \quad 
                 \textsuperscript{$\ddagger$}Corresponding Author.}
\begin{abstract}
Recent Multimodal Large Language Models (MLLMs) have significantly advanced e-commerce product understanding.
However,
they still face three challenges:
(i) the modality imbalance induced by modality mixed training;
(ii) underutilization of the intrinsic alignment relationships among visual and textual information within a product;
and (iii) limited handling of noise in e-commerce multimodal data.
To address these, we propose \model, 
a dynamic modality-balanced \textbf{M}ultim\textbf{O}dal representation learning framework for e-commerce pr\textbf{O}duct u\textbf{N}derstanding.
It comprises:
(1) a Modality-driven Mixture-of-Experts (MoE) that adaptively processes input samples by their modality composition, 
enabling Multimodal Joint Learning to mitigate the modality imbalance;
(2) a Dual-level Alignment method to better leverage semantic alignment properties inside individual products;
and (3) an MLLM-based Image-text Co-augmentation strategy that integrates textual enrichment with visual expansion, coupled with Dynamic Sample Filtering to improve training data quality.
We further release \bench, a co-augmented \textbf{M}ultimodal representation \textbf{B}enchmark for 
\textbf{E}-commerce representation learning and evaluation at \url{https://huggingface.co/datasets/ZHNie/MBE2.0}.
Experiments show that \model delivers state-of-the-art zero-shot performance on \bench and multiple public datasets. 
Furthermore, attention-based heatmap visualization provides qualitative evidence of improved multimodal alignment of \model.

\end{abstract}    
\section{Introduction}
\label{sec:intro}
\begin{figure}[t]
    \centering
    \includegraphics[width=1.0\linewidth]{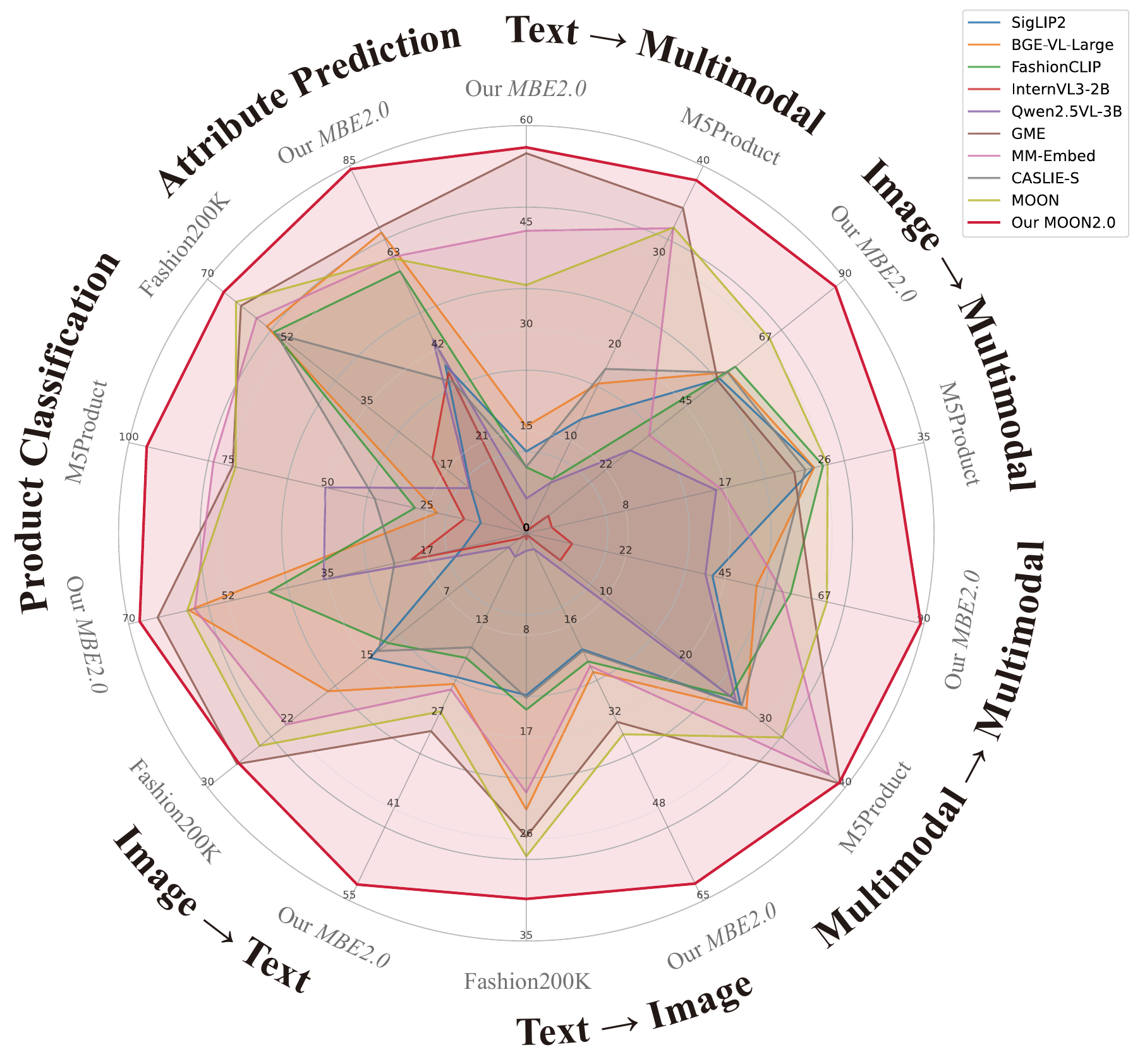}
    \caption{Overall results on all the downstream tasks. The arrows in the diagram indicate the target directions of the retrieval tasks.}
    \label{fig:radar}
\end{figure}

E-commerce has grown rapidly recently, 
driving the need for advances in product understanding tasks such as product retrieval~\cite{li2021embedding,hendriksen2022multimodal,zheng2023delving} and product recommendation~\cite{iqbal2018multimodal,liu2024multimodal}. 
To improve generalization across various tasks and reduce training cost, 
the e-commerce representation learning approaches that utilize multimodal information of products have gained increasing attention as \textit{task-agnostic} solutions for product understanding~\cite{guo2018multi,li2020adversarial,liu2022pretraining,yan2025mim}.


Most of these works adopt the dual-flow architecture~\cite{yu2022commercemm,jin2023learning,wang2023missrec,jiang2024mrse,dai2024uniembedding}, which comprises 
a visual encoder and a textual encoder, to map these
modalities into a shared latent space.
However, this architecture is inherently unsuitable for modeling
the many-to-one relationship typical in e-commerce scenarios, where multiple images
(\eg, the stock-keeping unit (SKU) images and creative images)
are associated with a shared product title.


Inspired by the ability of MLLMs to project heterogeneous inputs into a unified embedding space~\cite{zhang2025bridging,jiang2025vlmvec,li2024llava},
e-commerce MLLMs have successfully leveraged multiple images and texts, showing a deeper understanding of products through richer multimodal information~\cite{shi-etal-2025-llama,ling2025captions,peng2024ecellm,zhang2026moon,fu2025moon}.
Despite these advances, existing MLLMs for product understanding still face several limitations.

\begin{figure}[t]
  \centering
   \includegraphics[width=\linewidth]{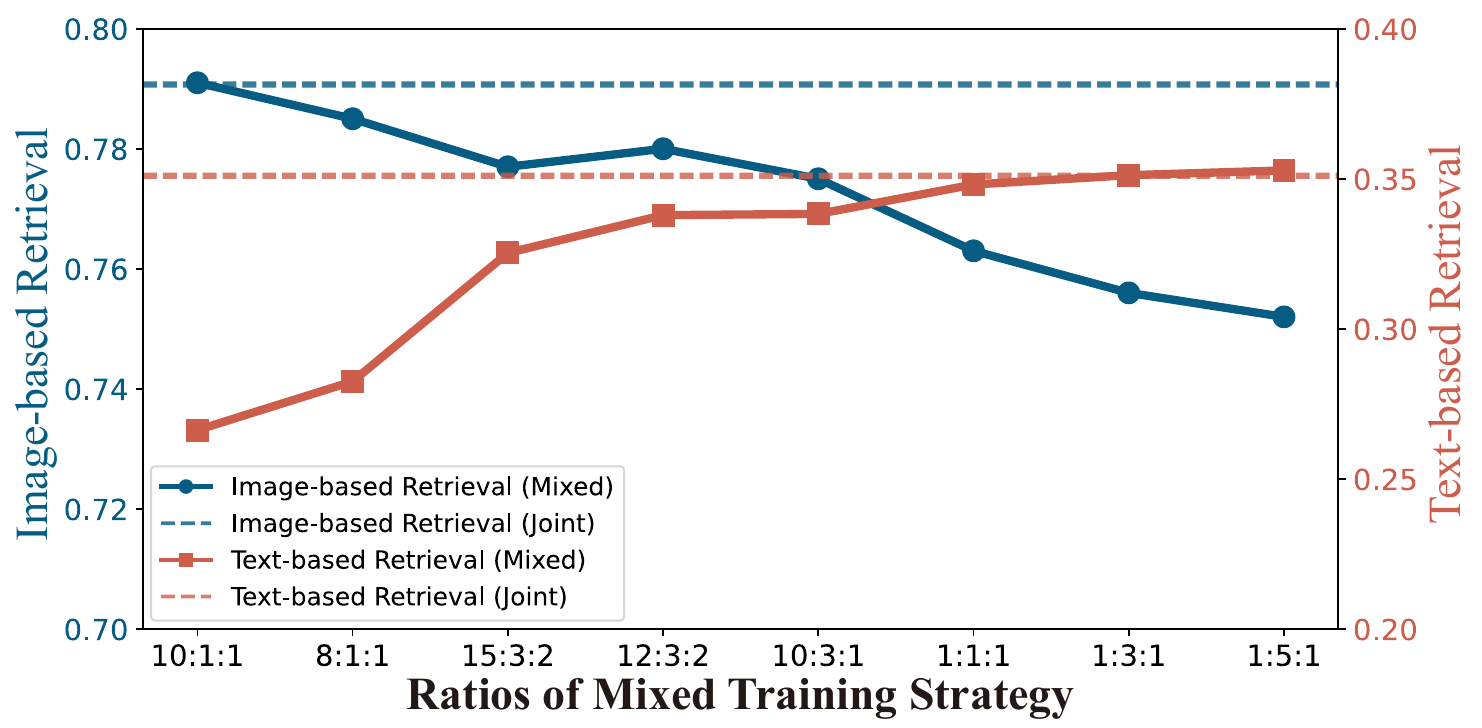}
   \caption{Modality imbalance under the training set of mixed training strategy. 
   The image-based and text-based retrieval refer to the image-to-multimodal and text-to-multimodal retrieval tasks,
   and the dashed line corresponds to the model performance following multi-objective joint training.}
   \label{fig:motivation}
\end{figure}

First, regarding training methods, they primarily employ modality mixed training with fixed ratios over modality combinations of 
queries, 
\eg, MOON~\cite{zhang2026moon} adopts a 12:3:2 mix of image-only, text-only, and multimodal queries containing both image and text, while positives always include both modalities.
However, as shown in \cref{fig:motivation}, this mixed strategy can induce \textit{modality imbalance}~\cite{zhang2024modality,fan2023pmr,chow2024unified}, leading to varying degradation on downstream tasks compared to multi-objective joint training, 
due to a mismatch between the fixed training mixture and the modality distribution of downstream tasks.

Second, in terms of supervision signals, existing methods for product representation learning mostly focus on the relationships between products (inter-product) in e-commerce 
scenarios~\cite{chia2022contrastive,yu2022commercemm,liu2023multimodal,tiady2024merlin}, with limited explicit modeling of modality information relationships within 
individual products (intra-product).
This leads to significantly reduced information utilization during training and fails to leverage the inherent semantic alignment properties of multimodal information inside individual products.

Finally, as for the data quality,
existing works primarily involve raw-data deduplication, category rebalancing, 
and rough core-product detection
~\cite{jiang2018mentornet,zhan2021product1m,dong2022m5product,ling2025ecommmmu,zhang2026moon}, 
without adequately performing noise mitigation and diversity expansion.
This limitation is pronounced in e-commerce, where texts often include redundant or noisy content, while images can be cluttered or lacking viewpoint diversity.


To address these issues, we propose \textbf{\model}, 
the first dynamic modality-balanced \textbf{M}ultim\textbf{O}dal representation learning framework for e-commerce pr\textbf{O}duct u\textbf{N}derstanding, 
which integrates the following techniques:

For model architecture and training, we incorporate a \textit{Modality-driven MoE} that dynamically routes experts to 
process input samples by their modality composition.
This design enables \textit{Multimodal Joint Learning}, which jointly optimizes contrastive objectives over image-only, text-only, and multimodal queries, aligning them with multimodal targets within a single end-to-end training stage.

For optimization objectives, we introduce a \textit{Dual-level Alignment} method to conduct alignment based on inter- and intra-product relationships.
In addition to the focus on the relationships between products (inter-product alignment), it explicitly aligns the image and text modalities within each product (intra-product learning), leveraging stronger image-text alignment semantics for product modeling.

For data augmentation, we propose an \textit{MLLM-based Image-text Co-augmentation} strategy that couples textual enrichment with multi-granularity visual expansion, including the extraction of main subject image, diversification of background, and refinement of logos and fine-grained details.
Moreover, \textit{Dynamic Sample Filtering} is integrated into the contrastive learning pipeline to reduce the impact of noise in the raw e-commerce data.

Furthermore, we release \textbf{\bench}, a co-augmented \textbf{M}ultimodal representation \textbf{B}enchmark for \textbf{E}-commerce to advance e-commerce representation learning.
It contains a co-augmented multimodal training set and a test set supporting various downstream tasks, including product retrieval, classification, and attribute prediction.
Extensive experiments on \bench and public datasets (\eg, M5Product~\cite{dong2022m5product}, Fashion200K~\cite{han2017automatic}) show that \model achieves state-of-the-art zero-shot performance across these tasks.
Furthermore, attention-based heatmap visualization on the e-commerce data of \bench provides qualitative evidence of improved multimodal alignment of \model.
Our main contributions are summarized as follows:

\begin{itemize}
    \item 
    We propose \model, the first dynamic modality-balanced multimodal representation learning framework for e-commerce product understanding. Built upon an
    MLLM, \model supports \textit{Multimodal Joint Learning} 
    to mitigate the modality imbalance.
    \item 
    For the aspect of model architecture, training strategies, and data augmentation, 
    we employ the \textit{Modality-driven MoE}, \textit{Dual-level Alignment}, \textit{MLLM-based Image-text Co-augmentation}, and \textit{Dynamic Sample Filtering}.
    Together, they effectively exploit inter- and intra-product relationships for product understanding.
    \item 
    We release \bench, a co-augmented multimodal representation benchmark with 6.4 million real-world e-commerce samples. 
    Extensive experiments on \bench and public datasets show that \model achieves SOTA performance across diverse downstream tasks, and attention-based heatmap visualization further provides qualitative evidence of improved image-text alignment.
\end{itemize}

\begin{figure*}[t]
  \centering
  \includegraphics[width=\linewidth]{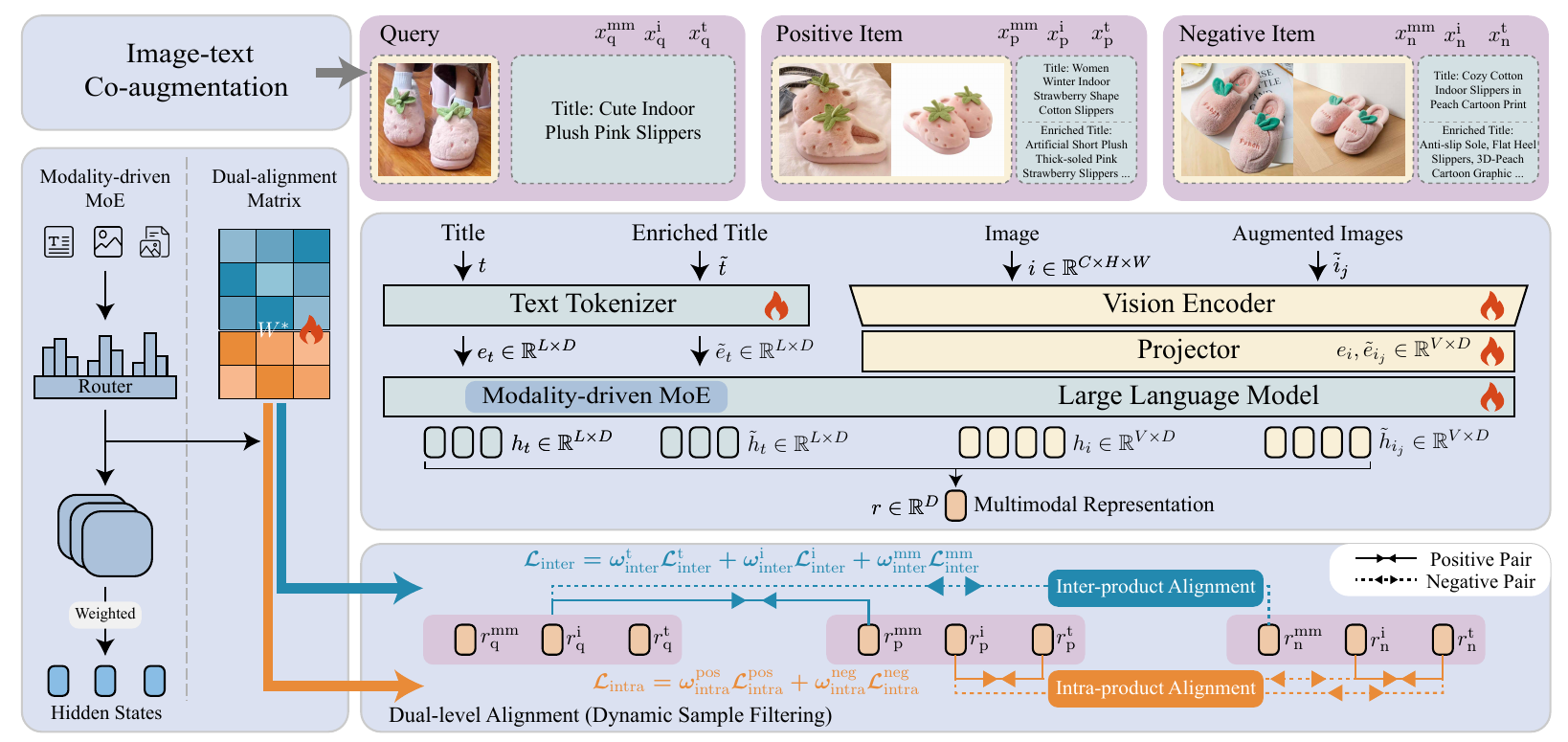}
  \caption{Pipeline of our \model.
  Given a training triplet consisting of a query, a positive item, and a negative item, the model processes each element into three input modalities: multimodal ($x^{mm}$, combining both image and text), image-only ($x^{i}$), and text-only ($x^{t}$). 
  In addition, positive item and negative item will further include enriched title and augmented images.
  }
  \label{fig:Pipeline}
\end{figure*}
\section{Related Work}
\label{sec:relatedwork}
\subsection{E-commerce Representation Learning}
Early e-commerce systems rely on dual-flow architectures with separate encoders aligned via similarity or retrieval objectives~\cite{gao2020fashionbert}. 
While unified contrastive frameworks~\cite{yu2022commercemm,chen2022product2vec,chia2022contrastive,dai2024uniembedding} and gated fusion methods~\cite{liang2025uniecs} improved upon large-scale backbones~\cite{radford2021learning,alayrac2022flamingo,li2022blip}, they struggle with many-to-one product relationships.
Inspired by the strong semantic capacity of MLLMs~\cite{lin2025mmembed,team2023gemini,zhang2025bridging,achiam2023gpt,bai2023qwen}, generative-model-based frameworks have begun to address this gap and offer richer multimodal representations~\cite{ling2025captions,zhang2026moon,ling2025ecommmmu}.
However, they are still limited by the modality imbalance.
Our \model framework aims to mitigate these issues by 
\textit{Multimodal Joint Learning} for efficient and modality-balanced multimodal representation learning.

\subsection{Multimodal Representation Learning}
Early multimodal approaches, including dual-flow models~\cite{radford2021learning,jia2021scaling,yuan2021florence} and fusion architectures~\cite{li2021align,li2022blip,alayrac2022flamingo}, achieve strong cross-modal alignment via contrastive learning and attention.
However, their rigid designs often overlook many-to-one relationships and limit adaptability across diverse tasks due to high computational costs.
While recent MLLMs~\citep{peng2024grounding,chen2024scaling,achiam2023gpt,team2023gemini,bai2023qwen,zhang2025sharper} unify heterogeneous inputs with broad capabilities, they frequently lack the domain-specific knowledge required for e-commerce. 
To bridge this gap, we propose \model, the first dynamic modality-balanced multimodal representation learning framework for e-commerce product understanding.

\section{Method}
\label{sec:method}
\subsection{Formulation and Overview}
\label{subsec:overview}
We primarily evaluate models on multimodal product retrieval (image-to-multimodal, text-to-multimodal, and multimodal-to-multimodal)~\cite{lin2025mmembed}, while assessing generalization via product classification and attribute prediction~\cite{zhang2026moon}.
To probe the alignment capability after applying \textit{Dual-level Alignment},
we incorporate both text-to-image and image-to-text retrieval tasks,
providing a comprehensive assessment of e-commerce product understanding.

\vpara{Multimodal Retrieval.}
For various multimodal
retrieval tasks, we denote $\mathcal{Q} = \{ q | q \in \{ q^\text{text},q^\text{image},q^\text{multimodal} \} \}$ and $\mathcal{C} = \{ c | c \in \{ c^\text{text},c^\text{image},c^\text{multimodal} \} \}$ as queries and the large-scale candidate set. The model aims to produce multimodal embeddings that align the query $q\in \mathcal{Q}$ and candidate $c \in \mathcal{C}$ in a unified feature space, which can be abstracted as $F(\cdot)$. 
Then we use the $\text{sim}(F(q), F(c))$ to measure the positive associations between query $q$ and candidate $c$.
\begin{equation}
    \dot{c} = \argmax_{c \in \mathcal{C}} \text{sim}(F(q), F(c)) ,
\end{equation}
By instantiating modalities for $q,c$, this formulation covers image-to-multimodal, text-to-multimodal, and multimodal-to-multimodal retrieval, as well as text-to-image and image-to-text retrieval for a more comprehensive assessment.


\vpara{Framework Overview.}
Addressing the challenges of modality imbalance, limited modeling of intra-product relations, and insufficient noise-handling, 
we propose \model, a dynamic modality-balanced multimodal representation learning framework, as illustrated in \cref{fig:Pipeline}.
First, to enhance data diversity and robustness,
we employ an \textit{MLLM-based Image-text Co-augmentation} strategy that enriches textual descriptions and 
expands visual content at multiple granularities.
The augmented data are organized into query, positive, and negative triplets and encoded by a generative MLLM backbone to obtain unified multimodal embeddings.
Second, to address the modality imbalance, we introduce \textit{Multimodal Joint Learning}
equipped with a \textit{Modality-driven Mixture-of-Experts (MoE)}. This module dynamically routes input samples to specialized experts based on their modality composition.
Finally, we optimize representations via \textit{Dual-level Alignment}, 
which jointly learns inter-product contrastive objectives and intra-product image-text alignment, 
further reinforced by \textit{Dynamic Sample Filtering} to ensure training robustness.

\subsection{Multimodal Joint Learning}
\label{subsec:learning}
\vpara{Multimodal Representation Obtaining.}
We follow prior generative-model-based approaches to obtain unified multimodal representations.
For each augmented triplet (query, positive, negative) constructed in \cref{sec:benchmark}, 
we instantiate multimodal, image-only, and text-only inputs for all three elements.
To unify encoding these three input modalities, we adopt a unified structure: image-only inputs are paired with an instructional prompt, whereas text-only inputs are processed directly in a single stream. 
Specifically, for the input text, the product title and enriched title are tokenized into text embeddings,
while the product image together with its augmentations is encoded by a vision encoder and projector into visual tokens.
Then the text and visual tokens are jointly processed by the LLM to produce a unified representation.
Notably, we adopt the \textit{Modality-driven MoE} for the feed-forward layers of the LLM, dynamically producing embeddings of input samples of diverse modality compositions.
Finally, we aggregate hidden states from the LLM's last layer by mean pooling to obtain a final representation.

In detail, the original text $t$ and enriched title $\tilde{t}$ are embedded as $e_{t}, \tilde{e}_{t} \in \mathbb{R}^{L\times D}$, where $L, D$ denote the feature length of text input and the hidden dimension of the LLM backbone.
The product image $i$ and $n_{c}$ augmented images $\{\tilde{i_{j}}, j \in [0,n_{c})\}$ are processed by vision encoder and projector into visual embeddings $e_{i}, \{\tilde{e}_{i_{j}}\} \in \mathbb{R}^{V\times D}$, where 
$V$ denotes the number of visual tokens per image.
Feeding $e_{t}, \tilde{e}_{t}, e_{i}$, and $\{\tilde{e}_{i_{j}}\}$ into the LLM yields hidden states $h \in \mathbb{R}^{(2L+(n_{c}+1)\times V)\times D}$ from last layer.
Finally, we apply mean pooling over h to obtain $r \in \mathbb{R}^{D}$, which serves as the representation for end-to-end \textit{Multimodal Joint Learning}.

\begin{figure}[t]
  \centering
   \includegraphics[width=\linewidth]{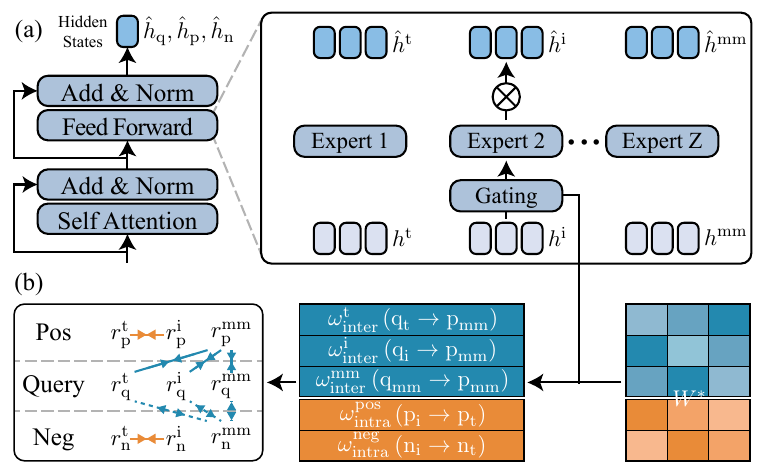}
   \caption{
   (a) \textit{Modality-driven MoE}. We adopt the MoE module for the feed-forward layers of the LLM backbone.
   Each of $\{\hat{h}_{\text{q}},\hat{h}_{\text{p}},\hat{h}_{\text{n}}\}$ includes hidden states for $\text{t},\text{i},\text{mm}$ input modalities.
   (b) \textit{Dual-level Alignment}. Besides inter-product alignment, we introduce intra-product alignment to further leverage the semantic consistency within the e-commerce products.
   The arrow symbol ($\rightarrow$) denotes the alignment relationship.
   Specifically, $\text{q},\text{p},\text{n}$ represent the query, positive item, and negative item, while $\text{t},\text{i},\text{mm}$ denote text-only, image-only, and multimodal input modalities.}
   \label{fig:moe}
\end{figure}

\vpara{Modality-driven MoE.}
Unlike conventional architectures that rely solely on token-level routing~\cite{shazeer2017outrageously, fedus2022switch} or constraint-based distillation without objective coupling~\cite{fang2025emoe}, 
we propose a \textit{Modality-driven Mixture-of-Experts (MoE)} module that 
integrates token-wise activation with modality-aware objective optimization.
We employ MoEs in the feed-forward layers of the LLM. 
Formally, given hidden states $h$, a gating network produces activations over experts:
$G = \text{softmax}(W_g h)$,
where $W_{g}$ denotes the gating linear layer.
Then we yield token-level expert outputs:
\begin{equation}
\hat{h} = \sum_{z=1}^{Z} \tilde{G}_{z} \cdot f_z(h),
\label{eq:moe_forward}
\end{equation}
where $f_{z}(\cdot)$ represents the $z$-th expert MLP, and $\tilde{G}_{z}$ denotes the normalized routing weight of the selected experts.

However, token-level routing alone is agnostic to 
the modality composition of inputs,
which can lead to suboptimal expert utilization across diverse alignment objectives
(\eg, $\text{q}_{\text{i}} \rightarrow \text{p}_{\text{mm}}$ for the alignment between the image-only query input and multimodal input of the positive item).
To introduce structured supervision, we define a learnable \textit{Dual-alignment Matrix} 
$W^{*} \in \mathbb{R}^{Z \times M}$, 
where $M$ is the number of 
alignment objectives. 
Each entry $W^{*}_{z, m}$ quantifies the intrinsic preference of expert $z$ for objective $m$. 
And $p_{z,m}$, the preference of expert $z$ over objective $m$, is normalized across objectives via softmax:
\begin{equation}
p_{z,m} = \frac{\exp(W^{*}_{z,m})}{\sum_{k=1}^{M} \exp(W^{*}_{z,k})}.
\label{eq:expert_pref}
\end{equation}
To couple expert specialization with diverse alignment objective optimization, we compute an objective-specific weight $\omega_m$ 
by aggregating token-level routing and learned expert preferences. 
Specifically, for each sample $b$ in the batch 
related to the query inputs that participate in the alignment objective $m$ (\eg, the text-only input of the query in the alignment
$\text{q}_{\text{t}} \rightarrow \text{p}_{\text{mm}}$),
let $\tilde{G}_{z,b}$ denote the routing weight of expert $z$ for that query representation. 
We define:
\begin{equation}
\omega_m = \frac{1}{\lvert\mathcal{B}_{m}\rvert} \sum_{b \in \mathcal{B}_{m}} \sum_{z=1}^{Z} p_{z,m} \cdot \tilde{G}_{z,b},
\label{eq:objective_weight}
\end{equation}
where $\mathcal{B}_{m}$ is the set of batch indices associated with objective $m$. 
Intuitively, $\omega_m$ reflects the collective expert support for optimizing the alignment objective $m$.

Beyond the standard auxiliary load-balancing term $\mathcal{L}_{\text{aux}}$ for even expert utilization~\cite{fedus2022switch},
we introduce a sparsity regularization loss $\mathcal{L}_{\text{sparsity}}$ to compel experts to specialize in specific alignment objectives. Specifically, we minimize the entropy of per-expert objective preference distributions:
\begin{equation}
\mathcal{L}_{\text{sparsity}} = 
\frac{1}{Z} \sum_{z=1}^{Z} \left( -\sum_{m=1}^{M}p_{z,m}\log p_{z,m}\right).
\label{eq:loss_sparsity}
\end{equation}
By reducing entropy, 
the preference of each expert over alignment objectives
is driven towards a more peaked (sparse) distribution, effectively guiding experts to specialize in a limited set of modality alignment objectives.

\vpara{Dual-level Alignment.}
To explicitly model both inter- and intra-product relationships, 
we formulate \textit{Dual-level Alignment} as a multi-objective contrastive training method that jointly optimizes inter-product alignment and intra-product alignment. 
Unlike dual-flow architectures~\cite{chia2022contrastive,bai2023cross},
this formulation aligns related products 
while preserving fine-grained semantic consistency between modalities within each product in many-to-one e-commerce contexts.

\vvpara{Inter-product Alignment.}
To align the representation of related products (inter-product),
we construct the triplet 
\((\text{q}, \text{p}, \text{n})\), 
denoting a query, a positive, and a negative.
For each triplet, 
we apply the following contrastive objective:
\begin{equation}
\mathcal{L}_{\text{inter}}^{\varphi} = - \log 
\frac{\exp \left( r_{\text{q}}^{\varphi}\cdot r_{\text{p}}^{\text{mm}} / \tau \right)}
     {\exp \left( r_{\text{q}}^{\varphi} \cdot r_{\text{p}}^{\text{mm}} / \tau \right) + 
      \sum_{\mathcal{N}_\text{q}} \exp \left( r_{\text{q}}^{\varphi}\cdot r_{\text{n}}^{\text{mm}} / \tau \right)},
\label{eq:inter_align}
\end{equation}
where $r_{\text{q}}^{\varphi} \ (\varphi\in \{\text{t}, \text{i}, \text{mm}\})$ is the representation of the text-only, image-only, or multimodal query,
and $r_{\text{p}}^{\text{mm}}, r_{\text{n}}^{\text{mm}}$ are the multimodal representations of the positive and negative items, 
and $\tau$ is a temperature.
The overall inter-product objective is
$\mathcal{L}_{\text{inter}}=\omega_{\text{inter}}^{\text{t}}\mathcal{L}_{\text{inter}}^{\text{t}}+\omega_{\text{inter}}^{\text{i}}\mathcal{L}_{\text{inter}}^{\text{i}}+\omega_{\text{inter}}^{\text{mm}}\mathcal{L}_{\text{inter}}^{\text{mm}}$, where the $\omega$ denotes the weight of each alignment objective.

\vvpara{Intra-product Alignment.}
While inter-product alignment conducts alignment between the related products,
we further introduce 
a finer-grained objective that strengthens image-text coherence within each product.
Given an image $i^{\psi}$ with its paired text $t^{\psi}$ and an unrelated text $t^{\psi^{-1}}$ extracted from the positive and negative pair of the triplet (q, p, n) mentioned before,
we define:
\begin{equation}
\mathcal{L}_{\text{intra}}^{\psi} = - \log 
\frac{\exp \left( r^\text{i}_{\psi}\cdot r^\text{t}_{\psi} / \tilde{\tau} \right)}
     {\exp \left( r^\text{i}_{\psi}\cdot r^\text{t}_{\psi} / \tilde{\tau} \right) + 
      \sum_{\text{t}^{\psi^{-1}}} \exp \left( r^\text{i}_{\psi}\cdot r^\text{t}_{\psi^{-1}} / \tilde{\tau} \right)},
\label{eq:intra_align}
\end{equation}
where $\tilde{\tau}$ is the intra-product temperature, and 
$\psi\in\{\text{pos}, \text{neg}\},\text{pos}^{-1}=\text{neg}$. 
The overall intra-product objective is
$\mathcal{L}_{\text{intra}}=\omega_{\text{intra}}^{\text{pos}}\mathcal{L}_{\text{intra}}^{\text{pos}}+\omega_{\text{intra}}^{\text{neg}}\mathcal{L}_{\text{intra}}^{\text{neg}}$.

\vpara{Holistic Optimization.}
The final optimization objective is
\begin{equation}
\mathcal{L}_{\text{total}}
= 
\mathcal{L}_{\text{inter}}
+\mathcal{L}_{\text{intra}}
+ \alpha \mathcal{L}_{\text{aux}}
+ \beta \mathcal{L}_{\text{sparsity}} ,
\label{eq:holistic_loss}
\end{equation}
where $\alpha,\beta$ weight the MoE regularization. 
It optimizes a unified embedding space for simultaneous inter- and intra-product coherence, promoting robust, modality-balanced multimodal representation learning.

\begin{figure}[t]
  \centering
   \includegraphics[width=1.0\linewidth]{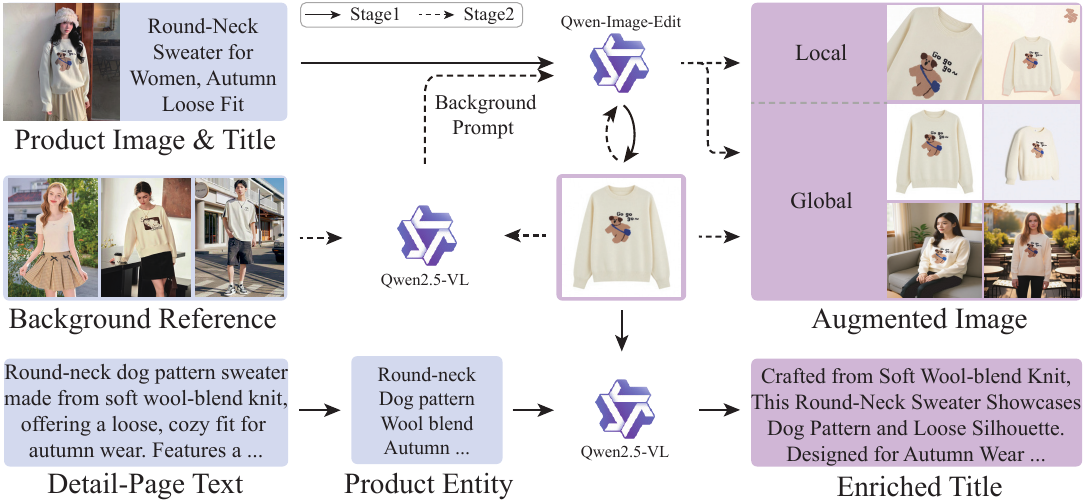}
   \caption{The \textit{MLLM-based Image-text Co-augmentation} pipeline.}
   \label{fig:augmentation}
\end{figure}

\begin{figure*}
  \centering
  \includegraphics[width=\linewidth]{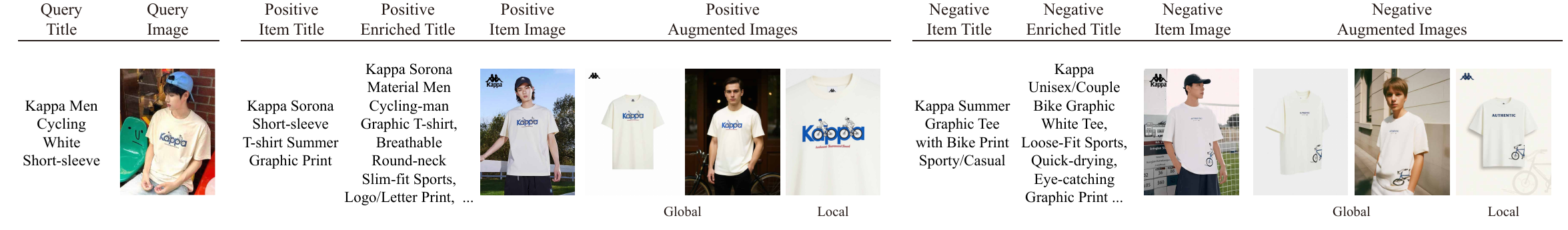}
  \caption{
    Illustration of one training sample in our \bench benchmark.
    Each training sample includes the title and image of the query, positive item, and negative item, with enriched titles and augmented images provided for the positive and negative items. 
    For evaluation, each test sample contains the original product content along with category labels and attribute annotations of the positive item. 
    No textual enrichment or visual augmentation is applied to the test set, facilitating a fair assessment of generalization.
  }
  \label{fig:benchmark}
\end{figure*}

\subsection{MLLM-based Image-text Co-augmentation}
\label{subsec:augmentation}
To further enhance the data quality and diversity,
we incorporate a \textit{MLLM-based Image-text Co-augmentation} strategy that enriches 
multimodal
inputs, as illustrated in \cref{fig:augmentation}.

\vpara{Textual Enrichment.}
In e-commerce scenarios, product titles are typically redundant and lack semantic richness. Although product description pages provide more detailed information, such metadata often suffers from substantial noise, leading to the omission of key descriptive entities.


To address this, we employ an MLLM that performs entity-aware textual enrichment.
Given an item title $T$, its associated description $D$, and the corresponding product image $I$, we first extract salient entity candidates $\mathcal{E} = \{ e_1, e_2, \ldots, e_m \}$ from $T$ and $D$ using an internal product entity extraction tool.
The MLLM then expands $T$ into an enriched representation $T^{+}$ by integrating contextual entities with visual cues from $I$ under controlled prompting:
\begin{equation}
T^{+} = \text{MLLM}_{\text{text}}(T, I, \mathcal{E}).
\end{equation}
The resulting enriched title provides more complete semantic coverage of product attributes and usage scenarios. 

\vpara{Visual Expansion.}
To complement text enrichment,
we apply an MLLM for multi-granularity visual-level augmentation, which performs global enhancement (\eg, main subject image extraction, background
diversification) and local enhancement (\eg, logo and fine-grained detail refinement), to enrich visual semantics and preserve key product cues.

Given a product image $I$, the MLLM performs a two-stage visual augmentation. First, it edits $I$ to remove irrelevant content while preserving essential product attributes, yielding standardized main subject images $I^{m}$.
Second, following~\cite{zhang2025bridging, tian2023stablerep, gu2024compodiff}, the model leverages $I^{m}$ to synthesize diverse yet semantically consistent variants $\{ I_{1}^{c}, I_{2}^{c}, \ldots, I_{n}^{c} \}$ via context-guided editing, varying in background, viewpoint, and details:
\begin{equation}
I^{c}_k = \text{MLLM}_{\text{edit}}(I^{m}, T, \text{prompt}_k),
\quad k \in \{1,\ldots, n\},
\end{equation}
For background variation, the MLLM takes background references and the main subject $I^{m}$ to generate context-aware prompts $\text{prompt}_k$ (\eg, scene style or lighting). These prompts then guide visually coherent background synthesis while strictly preserving product integrity.

This process broadens visual distribution while aligning with enriched textual semantics, thereby reinforcing multimodal invariance.
Finally, we employ CLIP to assess image-title consistency, filtering out low-quality samples to ensure data reliability.

\vpara{Dynamic Sample Filtering.}
After co-augmentation greatly enriches multimodal semantics,
the noise of raw e-commerce data (\eg mislabels, pseudo-positive and pseudo-negative samples) can still degrade contrastive alignment. 
To address this, we introduce \textit{Dynamic Sample Filtering}, which integrates into the contrastive learning pipeline, dynamically estimating triplet reliability during training.
For each triplet $(r_\text{q}, r_\text{p}, r_\text{n})$ of the inputs, the reliability weight $\phi$ is defined as:
\begin{equation}
\phi = \sigma\!\left(\kappa ((r_\text{q} \cdot r_\text{p}) - (r_\text{q} \cdot r_\text{n}) - \bar{\Delta}) \right),
\end{equation}
where $\sigma(\cdot)$ denotes the sigmoid function and $\kappa$ controls its sharpness. 
We fix the reliability threshold at $\delta=0.6$, while $\bar{\Delta}$ decays over training to transition the focus from high-confidence samples to challenging ones.
Triplets with $\phi < \delta$ are down-weighted in the loss, thereby suppressing pseudo-positive and pseudo-negative samples. 
By adapting reliability as the embedding space evolves, this mechanism enables end-to-end denoising, fostering a stable and increasingly discriminative contrastive learning process.
\section{The \bench Benchmark}
\label{sec:benchmark}
\begin{table*}[!t]
    \caption{Zero-shot results of the multimodal retrieval, product classification, and attribute prediction tasks on our \bench benchmark.}
    \label{tab:ourbench}
    \centering
    \setlength\tabcolsep{3pt}
    \renewcommand{\arraystretch}{1.0}  
\begin{adjustbox}{width=\textwidth,center}
\begin{tabular}{lrrrrrrrrrrrrrrrrrrrrrrr}
\toprule
\multicolumn{1}{c}{\multirow{3}{*}{\diagbox{Methods}{Metrics}}} & 
\multicolumn{15}{c}{Multimodal Retrieval}& 
\multicolumn{4}{c}{\multirow{2}{*}{Product Classification}} &
\multicolumn{4}{c}{\multirow{2}{*}{Attribute Prediction}} \\
\cmidrule(lr){2-16}

&
\multicolumn{3}{c}{$q^{\text{t}} \to c^{\text{mm}}$} & 
\multicolumn{3}{c}{$q^{\text{i}} \to c^{\text{mm}}$} & 
\multicolumn{3}{c}{$q^{\text{mm}} \to c^{\text{mm}}$} &
\multicolumn{3}{c}{$q^{\text{t}} \to c^{\text{i}}$} &
\multicolumn{3}{c}{$q^{\text{i}} \to c^{\text{t}}$} \\
\cmidrule(lr){2-4} \cmidrule(lr){5-7} \cmidrule(lr){8-10} \cmidrule(lr){11-13} \cmidrule(lr){14-16} \cmidrule(lr){17-20} \cmidrule(lr){21-24}
                  & \multicolumn{1}{c}{R@1} & \multicolumn{1}{c}{R@5} & \multicolumn{1}{c}{R@10} & \multicolumn{1}{c}{R@1} & \multicolumn{1}{c}{R@5} & \multicolumn{1}{c}{R@10} & \multicolumn{1}{c}{R@1} & \multicolumn{1}{c}{R@5} & \multicolumn{1}{c}{R@10} & \multicolumn{1}{c}{R@1} & \multicolumn{1}{c}{R@5} & \multicolumn{1}{c}{R@10} & \multicolumn{1}{c}{R@1} & \multicolumn{1}{c}{R@5} & \multicolumn{1}{c}{R@10} &
                  \multicolumn{1}{c}{Acc.} & \multicolumn{1}{c}{Prec.} & \multicolumn{1}{c}{Rec.} & \multicolumn{1}{c}{F1} &
                  \multicolumn{1}{c}{Acc.} & \multicolumn{1}{c}{Prec.} & \multicolumn{1}{c}{Rec.} & \multicolumn{1}{c}{F1}\\
\midrule
SigLIP2~\cite{tschannen2025siglip}  
& 4.94 & 12.05 & 16.23 
& 23.58 & 54.68 & 65.86 
& 18.00 & 42.13 & 52.17 
& 7.08 & 20.53 & 28.10 
& 8.34 & 21.54 & 28.64 
& 11.18 & 31.76 & 12.88 & 9.52 
& 38.91 & 53.86 & 44.08 & 37.08 \\
BGE-VL-Large~\cite{zhou2025megapairs}
& 8.34 & 15.76 & 18.92 
& 24.48 & 57.07 & 64.84 
& 24.80 & 52.12 & 63.33 
& 8.53 & 24.54 & 32.22 
& 9.30 & 22.57 & 30.24 
& 59.36 & \thr{62.37} & 57.37 & 54.03 
& \thr{69.61} & \two{79.51} & 67.11 & \thr{62.83} \\
FashionCLIP~\cite{chia2022contrastive}
& 4.02 & 9.73 & 12.95 
& \thr{26.50} & \thr{59.00} & \thr{69.20}
& 28.53 & 59.86 & 69.13 
& 8.22 & 22.67 & 28.89 
& 7.40 & 18.67 & 25.09 
& 45.27 & 56.25 & 48.75 & 42.89 
& 60.65 & 67.23 & 55.70 & 51.27 \\
\midrule
InternVL3-2B~\cite{zhu2025internvl3}  
& 0.11 & 0.37 & 0.61 
& 2.58 & 6.23 & 8.63 
& 4.76 & 10.38 & 13.45 
& 0.15 & 0.29 & 0.64 
& 0.18 & 0.31 & 0.76 
& 20.23 & 50.9 & 20.81 & 19.13 
& 37.31 & 78.10 & 41.68 & 37.97 \\
Qwen2.5-VL-3B~\cite{bai2025qwen2}  
& 1.81 & 5.13 & 7.58 
& 13.04 & 29.37 & 36.55 
& 18.50 & 40.57 & 48.77 
& 0.82 & 2.81 & 4.35 
& 1.19 & 3.51 & 5.42 
& 35.67 & 57.92 & 36.44 & 37.42 
& 44.83 & 69.46 & 45.07 & 40.38 \\
\midrule
GME~\cite{zhang2025bridging} 
& \two{27.20} & \two{55.93} & \one{64.41}
& 23.48 & 54.01 & 64.98 
& \thr{31.51} & \thr{64.57} & \thr{73.90} 
& \thr{14.30} & \thr{33.38} & \thr{41.77} 
& \thr{11.86} & \two{29.62} & \two{38.26} 
& \two{64.92} & \one{64.76} & \two{68.54} & \two{61.60} 
& \two{70.76} & 69.33 & \two{73.04} & \two{66.48} \\
MM-Embed~\cite{lin2025mmembed}  
& \thr{21.77} & \thr{44.52} & \thr{52.83} 
& 14.16 & 34.73 & 44.48 
& 28.45 & 58.01 & 66.67 
& 8.41 & 23.49 & 30.11 
& 8.77 & 23.40 & 31.56 
& 58.37 & 62.30 & \thr{64.25} & \thr{58.19} 
& 63.98 & \thr{78.79} & \thr{68.39} & 61.97 \\
CASLIE-S~\cite{ling2025captions} 
& 3.94 & 9.80 & 13.56 
& 25.73 & 56.90 & 67.02 
& 26.32 & 56.16 & 65.59 
& 8.16 & 20.81 & 28.16 
& 7.16 & 17.08 & 27.15 
& 23.25 & 38.93 & 23.99 & 18.22 
& 34.99 & 48.29 & 38.83 & 33.15\\
MOON~\cite{zhang2026moon} 
& 16.94 & 36.54 & 43.24 
& \two{32.18} & \two{68.78} & \two{78.11} 
& \two{35.18} & \two{68.06} & \two{80.78} 
& \two{17.79} & \two{35.56} & \two{44.02} 
& \two{12.56} & \thr{26.69} & \thr{36.65} 
& \thr{59.70} & 54.21 & 64.23 & 56.43 
& 63.55 & 36.57 & 40.00 & 36.95 \\
\midrule
Our \model  
& \one{27.34} & \one{56.80} & \two{63.09} 
& \one{41.07} & \one{87.31} & \one{91.08} 
& \one{43.34} & \one{89.31} & \one{94.21} 
& \one{25.05} & \one{61.99} & \one{73.12} 
& \one{20.47} & \one{52.59} & \one{64.91} 
& \one{68.08} & \two{63.88} & \one{73.60} & \one{65.68} 
& \one{84.29} & \one{82.07} & \one{82.99} & \one{79.39} \\
\bottomrule
\end{tabular}
\end{adjustbox}
\end{table*}
We construct our dataset from user logs on a leading Chinese e-commerce platform, Taobao, covering the period from January 1, 2023, to June 30, 2025.
Positive interactions are defined as purchases following a query, while skipped exposures with low relevance scores serve as negatives.
We first form (image query, positive item, negative item) triplets and (text query, positive item) pairs from image and text search logs, respectively.
Then, by joining on the positive item, we pair the text query with its corresponding image query to form a multimodal query (title and image), yielding complete triplets of (multimodal query, positive item, negative item).
All user data is anonymized, retaining only visual and textual product content.
The resulting dataset contains 5,751,594 training samples and 
636,241 test samples.

As detailed in \cref{subsec:augmentation}, 
we apply co-augmentation to the training set, enriching titles and expanding visual content at multiple granularities.
An overview of the training data format is illustrated in \cref{fig:benchmark}.
\section{Experiment}
\label{sec:experiment}
\begin{table}[!t]
    \caption{Zero-shot results of the multimodal retrieval and product classification tasks on M5Product benchmark.}
    \label{tab:M5Product}
    \centering
    \setlength\tabcolsep{3pt}
    \renewcommand{\arraystretch}{1.0}  
\begin{adjustbox}{width=\linewidth,center}
\begin{tabular}{lrrrrrrrrrc}
\toprule
\multicolumn{1}{c}{\multirow{3}{*}{\diagbox{Methods}{Metrics}}} & 
\multicolumn{9}{c}{Multimodal Retrieval}& 
\multicolumn{1}{c}{\multirow{2}{*}{Classification}}\\
\cmidrule(lr){2-10}

&
\multicolumn{3}{c}{$q^{\text{t}} \to c^{\text{mm}}$} & 
\multicolumn{3}{c}{$q^{\text{i}} \to c^{\text{mm}}$} & 
\multicolumn{3}{c}{$q^{\text{mm}} \to c^{\text{mm}}$} \\
\cmidrule(lr){2-4} \cmidrule(lr){5-7} \cmidrule(lr){8-10} \cmidrule(lr){11-11} 
                  & \multicolumn{1}{c}{R@1} & \multicolumn{1}{c}{R@5} & \multicolumn{1}{c}{R@10} 
                  & \multicolumn{1}{c}{R@1} & \multicolumn{1}{c}{R@5} & \multicolumn{1}{c}{R@10} 
                  & \multicolumn{1}{c}{R@1} & \multicolumn{1}{c}{R@5} & \multicolumn{1}{c}{R@10} 
                  & \multicolumn{1}{c}{Acc.}\\
\midrule
SigLIP2~\cite{tschannen2025siglip}  
& 2.41 & 8.49 & 12.44 
& 8.09 & 19.48 & 25.29 
& 9.53 & 21.71 & 26.87 
& 11.46 \\
BGE-VL-Large~\cite{zhou2025megapairs} 
& 4.04 & 11.17 & 16.31 
& \thr{8.94} & 19.43 & 25.34 
& 10.08 & 22.35 & 27.62 
& 22.42 \\
FashionCLIP~\cite{chia2022contrastive} 
& 1.16 & 2.53 & 5.88 
& 8.23 & 19.44 & \thr{26.13} 
& 9.21 & 20.76 & 25.65 
& 28.02 \\
\midrule
InternVL3-2B~\cite{zhu2025internvl3}  
& 0.13 & 0.25 & 0.38 
& 0.53 & 1.51 & 2.25 
& 1.19 & 2.97 & 4.26 
& 15.64 \\
Qwen2.5-VL-3B~\cite{bai2025qwen2} 
& 1.14 & 3.59 & 5.52 
& 4.88 & 12.31 & 16.72 
& 8.01 & 19.84 & 26.29 
& 50.54 \\
\midrule
GME~\cite{zhang2025bridging} 
& \two{13.96} & \two{24.25} & \two{35.42} 
& 6.66 & 17.87 & 23.60 
& \two{14.49} & \two{26.61} & \one{39.40} 
& \thr{73.87} \\
MM-Embed~\cite{lin2025mmembed} 
& \thr{11.90} & \thr{22.36} & 33.23 
& 5.78 & 13.57 & 17.14 
& 13.43 & \thr{25.31} & \thr{37.96} 
& \two{78.70} \\
CASLIE-S~\cite{ling2025captions} 
& 5.69 & 11.13 & 17.90 
& 8.00 & \thr{19.50} & 24.57 
& 8.40 & 20.44 & 27.04 
& 38.16 \\
MOON~\cite{zhang2026moon} 
& 10.14 & 20.33 & \thr{33.31} 
& \two{9.11} & \two{21.59} & \two{26.53} 
& \thr{14.27} & 25.05 & 32.13 
& 73.12 \\
\midrule
Our \model  
& \one{15.27} & \one{25.69} & \one{38.45} 
& \one{11.28} & \one{24.43} & \one{32.37} 
& \one{15.21} & \one{27.37} & \two{39.27} 
& \one{95.50} \\
\bottomrule
\end{tabular}
\end{adjustbox}
\end{table}

\begin{table}[t]
    \caption{Zero-shot results of the multimodal retrieval and product classification tasks on Fashion200K benchmark.}
    \label{tab:fashion200k}
    \centering
    \setlength\tabcolsep{3pt}
    \renewcommand{\arraystretch}{1.0}  
\begin{adjustbox}{width=\linewidth,center}
\begin{tabular}{lrrrrrrrrrr}
\toprule
\multicolumn{1}{c}{\multirow{3}{*}{\diagbox{Methods}{Metrics}}} & 
\multicolumn{6}{c}{Multimodal Retrieval}& 
\multicolumn{4}{c}{\multirow{2}{*}{Product Classification}}\\
\cmidrule(lr){2-7}

&
\multicolumn{3}{c}{$q^{\text{t}} \to c^{\text{i}}$} &
\multicolumn{3}{c}{$q^{\text{i}} \to c^{\text{t}}$} \\
\cmidrule(lr){2-4} \cmidrule(lr){5-7} \cmidrule(lr){8-11} 
                  & \multicolumn{1}{c}{R@1} 
                  & \multicolumn{1}{c}{R@5} 
                  & \multicolumn{1}{c}{R10} 
                  & \multicolumn{1}{c}{R@1} 
                  & \multicolumn{1}{c}{R@5} 
                  & \multicolumn{1}{c}{R10} 
                  & \multicolumn{1}{c}{Acc.} 
                  & \multicolumn{1}{c}{Prec.} 
                  & \multicolumn{1}{c}{Rec.} 
                  & \multicolumn{1}{c}{F1} \\
\midrule
SigLIP2~\cite{tschannen2025siglip}   
& 4.19 & 10.03 & 13.90 
& 4.39 & 10.49 & 14.70 
& 12.18 & 20.04 & 10.92 & 6.60 \\
BGE-VL-Large~\cite{zhou2025megapairs}  
& 6.75 & 18.62 & 23.71
& 7.27 & 14.21 & 18.65
& 56.97 & \two{67.40} & 62.12 & 55.31 \\
FashionCLIP~\cite{chia2022contrastive}  
& 4.92 & 11.67 & 15.14
& 4.34 & 9.15 & 12.97
& 55.42 & 65.39 & 63.13 & 54.66 \\
\midrule
InternVL3-2B~\cite{zhu2025internvl3}   
& 0.09 & 0.27 & 0.55 
& 0.12 & 0.11 & 0.62
& 20.59 & 29.16 & 19.61 & 18.00 \\
Qwen2.5-VL-3B~\cite{bai2025qwen2}    
& 0.45 & 1.06 & 1.50 
& 0.54 & 1.17 & 1.63
& 12.44 & 25.60 & 13.16 & 12.14\\
\midrule
GME~\cite{zhang2025bridging}    
& \thr{8.13} & \thr{20.31} & \thr{26.06}
& \thr{10.31} & \two{21.14} & \one{27.23}
& \thr{62.66} & 63.44 & \two{68.15} & \two{61.38} \\
MM-Embed~\cite{lin2025mmembed}   
& 5.97 & 15.71 & 22.25 
& 6.12 & 15.91 & 22.57 
& 59.24 & \thr{66.29} & 66.60 & 59.24 \\
CASLIE-S~\cite{ling2025captions}   
& 4.71 & 11.25 & 14.12
& 4.41 & 10.04 & 13.89
& 54.88 & 58.23 & 55.65 & 53.76 \\
MOON~\cite{zhang2026moon}   
& \two{10.82} & \two{22.89} & \two{27.73}
& \two{11.71} & \thr{20.02} & \thr{25.09}
& \two{63.74} & 59.02 & \thr{68.00} & \thr{61.05} \\
\midrule
Our \model   
& \one{13.05} & \one{25.25} & \one{31.39} 
& \one{13.10} & \one{23.16} & \two{27.09} 
& \one{66.44} & \one{68.90} & \one{69.55} & \one{64.21} \\
\bottomrule
\end{tabular}
\end{adjustbox}
\end{table}

\subsection{Experimental Setup}
\label{subsec:setup}
\vpara{Training.}
Based on our in-house developed generative-model-based MLLM for e-commerce,
we conduct single-stage supervised finetuning on our proposed training set detailed in \cref{sec:benchmark}.
The model was optimized using a learning rate of 1 $\times$ 10\textsuperscript{-5} with a cosine scheduler.
The entire experiment was trained for approximately 18 hours
on 64 GPUs (NVIDIA A100), with a batch size of 4 per GPU.

\vpara{Baselines.}
To better validate the product understanding capabilities of \model, we conduct comparisons with multimodal representation learning works. 
First, we conduct evaluations on the large-scale contrastive learning method Siglip2~\cite{tschannen2025siglip} and BGE-VL-Large~\cite{zhou2025megapairs}, as well as recent universal multimodal retrieval methods GME~\cite{zhang2025bridging} and MM-Embed~\cite{lin2025mmembed}.
To further compare with open-source MLLM on product understanding tasks, we also include InternVL3-2B~\cite{zhu2025internvl3} and Qwen2.5-VL-3B~\cite{bai2025qwen2}.
Finally, we include domain-specific models trained in e-commerce scenarios, a dual-flow model FashionCLIP~\cite{chia2022contrastive}, 
and generative-model-based methods CASLIE-S~\cite{ling2025captions} and MOON~\cite{zhang2026moon}.
We follow official implementations: dual-flow models (\eg, FashionCLIP) use separate encoders with $\ell_2$-normalized embeddings; generative MLLMs (\eg, Qwen2.5-VL, GME) apply mean-pooling over last-layer hidden states with a unified prompt ``Describe this product based on the image and title''. All methods share identical candidate sets and pooling strategies.
Furthermore, for equitable comparison, Qwen2.5-VL and InternVL3 were fine-tuned on the \bench training set, and the results can be found in the supplementary materials.

\vpara{Evaluation Tasks.}
Our model aims to achieve more comprehensive product understanding in e-commerce scenarios and deliver robust performance across multiple downstream tasks, solving the issue of modality imbalance.
Therefore, we evaluated our model on \bench, M5Product~\cite{dong2022m5product}, and Fashion200K~\cite{han2017automatic} benchmarks upon multimodal retrieval and product classification tasks.
And the attribute prediction task is further conducted on our \bench benchmark.
To fairly demonstrate the effectiveness of representation learning across datasets, all experiments were conducted in a zero-shot setting.
Specifically, for multimodal retrieval tasks, we adopt Recall@$k$ as the evaluation metric, which measures the probability that the ground-truth item appears among the top-$k$ results in the ranked list.
For the product classification and attribute prediction tasks, standard metrics, \ie, accuracy, precision, recall, and F1, are adopted to evaluate the model performance on these tasks.

\subsection{Experimental Results}
\label{subsec:results}
Results of \model and baselines across evaluation tasks are demonstrated in \cref{fig:radar}, highlighting state-of-the-art performance and the effectiveness of our architecture, training method, and data enhancement in e-commerce scenarios.
Detailed comparisons are discussed below. 
In the tables, 
we mark the first, second, and third entries in each column with (\textbf{v}), (\underline{v}), and (*v), respectively.
The arrows ($\to$) denote retrieval directions, where $q,c$ indicate query and candidate, and $\text{mm},\text{i},\text{t}$ denote multimodal, image, and text modalities.

The results on \bench benchmark are shown in \cref{tab:ourbench}.
Our \model attains leading results on most multimodal retrieval settings across different values of $k$.
Compared with dual-flow architectures such as SigLIP2 and FashionCLIP, \model demonstrates comprehensive performance gains, suggesting benefits from the many-to-one modeling capacity of generative-model-based MLLMs.
Besides these baselines, \model also outperforms MOON, CASILE, and other MLLM-based retrieval models, highlighting its ability to process and align
heterogeneous query-target pairs of diverse modality composition
enabled by the \textit{Modality-driven MoE} for \textit{Multimodal Joint Learning}.
Notably, \model delivers larger improvements on non-traditional cross-modal settings (\eg, $q^{\text{t}} \to c^{\text{i}}$ and $q^{\text{i}} \to c^{\text{t}}$), which we attribute to \textit{Dual-level Alignment} within the model, which effectively leverages explicit semantic relationships between products in e-commerce data and image-text alignment semantics inside individual products.

For product classification and attribute prediction, \model also achieves SOTA performance. 
BGE-VL-Large, GME, and MM-Embed also perform strongly, 
highlighting their focus on classification and clustering.
Although these models show competitive zero-shot precision, we note that accuracy and F1 provide more holistic evaluation metrics by balancing precision and recall.
Compared to e-commerce-specific models such as FashionCLIP and MOON, 
\model benefits from the \textit{Multimodal Joint Learning} and \textit{MLLM-based Image-text Co-augmentation}, which provide more informative and better-aligned signals, yielding consistent gains on these tasks.


\cref{tab:M5Product,tab:fashion200k} show the evaluation on the M5Product and Fashion200K benchmark. 
\model demonstrates competitive performance on both datasets, further validating its effectiveness.
These findings confirm the robustness and generality of the proposed improvements across diverse evaluation tasks and data distributions.

\begin{table}[!t]
    \caption{Ablation study on our \bench benchmark.}
    \label{tab:ablation}
    \centering
    \setlength\tabcolsep{3pt}
    \renewcommand{\arraystretch}{1.0}  
\begin{adjustbox}{width=\linewidth,center}
\begin{tabular}{lccccccc}
\toprule
\multicolumn{1}{c}{\multirow{2}{*}{\diagbox{Ablation}{Metrics}}} & 
\multicolumn{5}{c}{Retrieval (R@10)}& 
\multicolumn{1}{c}{Class.}&
\multicolumn{1}{c}{Attr.}\\
\cmidrule(lr){2-6} \cmidrule(lr){7-7} \cmidrule(lr){8-8}

&$q^{\text{t}} \to c^{\text{mm}}$
&$q^{\text{i}} \to c^{\text{mm}}$
&$q^{\text{mm}} \to c^{\text{mm}}$
&$q^{\text{t}} \to c^{\text{i}}$
&$q^{\text{i}} \to c^{\text{t}}$
&Acc.
&Acc.\\

\midrule
Our \model 
& \one{63.09} & \one{91.08} & \one{94.21} & \one{73.12} & \one{64.91} & \one{68.08} & \one{84.29} \\
$w/o$ MoE 
& 51.29 & 74.59 & 78.45 & 62.16 & 56.21 & 62.55 & 75.62 \\
$w/o$ Alignment  
& 37.99 & 65.72 & 67.45 & 31.45 & 23.35 & 57.12 & 67.24 \\
$w/o$ Co-augmentation 
& 59.69 & 78.17 & 80.62 & 64.79 & 58.68 & 66.21 & 77.77 \\
$w/o$ Filtering 
& 60.63 & 83.40 & 80.00 & 70.40 & 63.21 & 67.99 & 84.04 \\
\bottomrule
\end{tabular}
\end{adjustbox}
\end{table}


\subsection{Ablation Study}
\label{subsec:ablation}
To quantify the contribution of the components in our \model, we conduct ablations on four variants: 
(1) $w/o$ \textit{Modality-driven MoE}, 
(2) $w/o$ \textit{Dual-level Alignment}, 
(3) $w/o$ \textit{MLLM-based Image-text Co-augmentation}, 
(4) $w/o$ \textit{Dynamic Sample Filtering}. 
Results across key tasks are summarized in \cref{tab:ablation}.
Overall, \model remains strong and stable, while removing individual components yields measurable declines.
Removing the \textit{Modality-driven MoE} leads to degradation across settings, as replacing experts with a plain MLP is less suited to heterogeneous query-target pairs, limiting specialization of embedding production and reducing alignment quality. More analysis of the effectiveness of Modality-driven MoE can be found in the supplementary materials.
Without \textit{Dual-level Alignment}, performance drops most notably on tasks that probe multimodal alignment, and retrieval metrics become less balanced. This indicates that, while inter-product retrieval patterns may still be learned in other training methods, the absence of intra-product constraints weakens semantic and modality coherence. 
Finally, eliminating \textit{MLLM-based Image-text Co-augmentation} or \textit{Dynamic Sample Filtering} produces modest declines on most downstream tasks: without co-augmentation, inputs are less informative; without dynamic filtering, training becomes more sensitive to the noise of raw e-commerce data.

\subsection{Heatmap Visualization}
\label{subsec:visualization}
\begin{figure}[t]
  \centering
   \includegraphics[width=\linewidth]{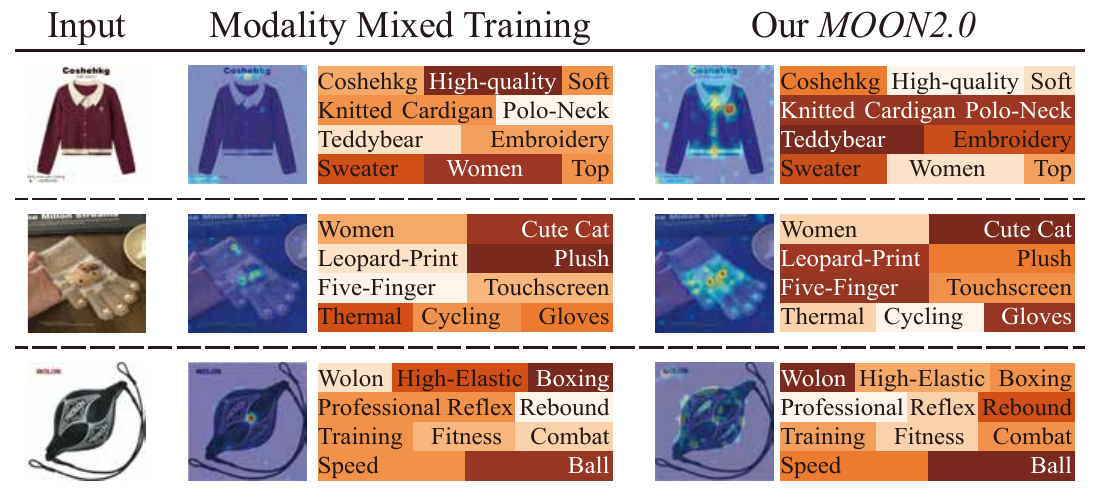}
   \caption{The heatmap visualization of our \model.}
   \label{fig:visualization}
\end{figure}

A set of attention-based heatmap visualizations is shown in \cref{fig:visualization}.
Comparison of visualization results on models using \model and using modality mixed training, clearly demonstrates that \model comprehensively improves the understanding capability of multimodal content and finer image-text alignment.
Specifically, it effectively attends to the main products while achieving finer alignment on detailed regions. For example, in the first row, attention shifts from non-critical regions such as ``high quality" and ``women" to focus on key attributes like ``knitted cardigan", ``polo-neck", and ``Teddybear", while also highlighting the brand name ``Coshehkg".

\section{Conclusion}
\label{sec:conclusion}
Our \model successfully achieves dynamic modality-balanced multimodal representation learning in the e-commerce domain by enhancing the model architecture, training method, and data augmentation strategies, effectively addressing the modality imbalance that plagued previous approaches.
The proposed architecture and training paradigm suggest promising directions for broader applications, which we believe can further advance general-purpose multimodal representation learning while maintaining efficiency and robustness.

{
    \small
    \bibliographystyle{ieeenat_fullname}
    \bibliography{main}
}

\end{document}